\documentclass{article}

\usepackage[dblblindworkshop, final]{neurips_2026}
\workshoptitle{Foundation Models for Temporal Systems: From Forecasting to World Modeling}

\usepackage[utf8]{inputenc} %
\usepackage[T1]{fontenc}    %
\usepackage{hyperref}       %
\usepackage{url}            %
\usepackage{booktabs}       %
\usepackage{amsmath,amsthm,amsfonts}
\usepackage{nicefrac}       %
\usepackage{microtype}      %
\usepackage{xcolor}         %
\usepackage{graphicx}
\usepackage{tabularx}
\usepackage{makecell}
\usepackage{ragged2e}
\usepackage{multirow}
\usepackage{pifont}
\usepackage{xspace}
\usepackage{placeins}
\usepackage{wrapfig}

\newcolumntype{Y}{>{\RaggedRight\arraybackslash}X}

\newcommand{\model}[1]{\texttt{#1}\xspace}
\newcommand{\ChronosCov}{\model{Chronos-2}}
\newcommand{\TimesFMCov}{\model{TimesFM-2.5}}
\newcommand{\MoiraiCov}{\model{Moirai-1.1}}
\newcommand{\ApolloPFN}{\model{ApolloPFN}}
\newcommand{\TabPFNvTS}{\model{TabPFN-TS}}
\newcommand{\TabICL}{\model{TabICL-v2}}
\newcommand{\ChronosUni}{\model{Chronos}}
\newcommand{\TimesFMUni}{\model{TimesFM}}

\newcommand{\CRN}{\model{CRN}}
\newcommand{\CT}{\model{CT}}
\newcommand{\IGCNet}{\model{IGC-Net}}
\newcommand{\RMSN}{\model{RMSN}}
\newcommand{\DoFlow}{\model{DoFlow}}
\newcommand{\CaTSG}{\model{CaTSG}}
\newcommand{\DoPFN}{\model{Do-PFN}}
\newcommand{\CausalPFN}{\model{CausalPFN}}
\newcommand{\CausalFM}{\model{CausalFM}}
\newcommand{\CausalTimePrior}{\model{CausalTimePrior}}
\newcommand{\CaML}{\model{CaML}}

\newcommand{\scenInDist}{\textsc{In-Distribution}\xspace}
\newcommand{\scenOOS}{\textsc{Out-of-Support}\xspace}
\newcommand{\scenOnset}{\textsc{Policy-Onset}\xspace}
\newcommand{\scenSwitch}{\textsc{Policy-Switch}\xspace}

\newcommand{\cmark}{\ding{51}}
\newcommand{\xmark}{\ding{55}}

\newcommand{\dataset}[1]{\texttt{#1}\xspace}
\newcommand{\Mthree}{\dataset{M3}}
\newcommand{\Mfour}{\dataset{M4}}
\newcommand{\Mfive}{\dataset{M5}}
\newcommand{\CovNovBench}{\dataset{CovNovBench}}
\newcommand{\FevBench}{\dataset{fev-bench}}
\newcommand{\GIFTEval}{\dataset{GIFT-Eval}}
\newcommand{\MonashArchive}{\dataset{Monash}}
\newcommand{\CausalRivers}{\dataset{CausalRivers}}
\newcommand{\MIMICReal}{\dataset{MIMIC3-Real}}
\newcommand{\MIMICSynth}{\dataset{MIMIC3-Synth}}
\newcommand{\CancerTreatment}{\dataset{CancerTreatment}}

\newcommand{\eps}{\varepsilon}
\newcommand{\R}{\ensuremath{\mathbb{R}}}
\newcommand{\mc}[1]{\ensuremath{\mathcal{#1}}}
\newcommand{\CondP}[2]{\mathbb{P}\hspace{-0.5mm}\left\{#1 \hspace{0.1mm}\bigm|\hspace{0.1mm} #2\right\}}
\newcommand{\p}[1]{\ensuremath{\left(#1\right)}}
\newcommand{\sbr}[1]{\ensuremath{\left[#1\right]}}
\newcommand{\abs}[1]{\ensuremath{\left|#1\right|}}

\providecommand{\appendixref}[1]{Appendix~\ref{#1}}

\title{\CovNovBench: A Benchmark for Evaluating Time Series Foundation Models under Covariate Novelty}

\author{%
  Wenhao Pan\thanks{Work done during internship at Amazon Science} \\
  University of Washington \\
  Seattle, USA \\
  \And
  Ravi Kiran Selvam\thanks{Correspondence to ravisel@amazon.com.} \\
  Amazon Science \\
  Bellevue, USA \\ %
  \And
  Michael W.~Mahoney \\
  Amazon Science \\
  Berkeley, USA \\
  \And
  Mayank Jauhari \\
  Amazon Science \\
  Bellevue, USA \\
  \And
  Mengfei Cao\\
  Amazon Science \\
  Bellevue, USA \\
  \And
  Dmitry Efimov \\
  Amazon Science \\
  New York City, USA \\
  \And
  Boris N. Oreshkin \\
  Amazon Science \\
  Montreal, Canada \\
}

\hypersetup{
  pdftitle={CovNovBench: A Benchmark for Evaluating Time Series Foundation Models under Covariate Novelty},
  pdfauthor={Wenhao Pan, Ravi Kiran Selvam, Michael W. Mahoney, Mayank Jauhari, Mengfei Cao, Dmitry Efimov, Boris N. Oreshkin}
}

\begin{document}

\maketitle

\begin{abstract}
Covariate-aware time-series foundation models (TSFMs) condition their forecasts on additional input variables, called covariates, that are supplied over the future horizon.
This capability makes them appealing for interventional forecasting, in which a target is predicted under a planned action such as a new product pricing policy or a new patient treatment schedule.
We refer to covariates that encode such actions as treatment covariates.
Current covariate-aware TSFMs, however, are trained on observed patterns, and their behavior is unclear when future treatment covariates fall outside the range seen in the history.
Examples include a numerical treatment that leaves its historical range and a categorical treatment that takes a previously unseen policy.
We call this practically common setting covariate novelty, and note that it is not systematically represented in existing benchmarks.
We introduce an operational definition of covariate novelty and use it to construct \textbf{CovNovBench}, a benchmark of 88,476 evaluation series generated or filtered from four simulator-based and real-world datasets.
The benchmark spans categorical and numerical treatment covariates as well as stable and trending targets.
We evaluate six covariate-aware TSFMs zero-shot using point and directional accuracy metrics.
No model is consistently reliable in our benchmark: point error almost always increases under covariate novelty, directional performance changes non-uniformly, and the best observed model varies across datasets and metrics.
\CovNovBench provides a systematic testbed for developing TSFMs that forecast more reliably under covariate novelty.
\end{abstract}

\section{Introduction}
\label{sec:introduction}

Early time series foundation models (TSFMs) predict a horizon-$H$ future target path $Y_{T+1:T+H}$ from its length-$T$ history $Y_{1:T}$~\citep{ansari2024chronos,das2024timesfm}.
Recent TSFMs also condition on \emph{covariates}, time-indexed inputs supplied over the history, forecast horizon, or both~\citep{ansari2025chronos, potapczynski2026time}.
When future covariates encode planned actions such as prices or treatments, these models provide a zero-shot interface for assessing actions before they are taken.
If such a path is actively set rather than observed, predicting the target is referred to as \emph{interventional forecasting}~\citep{schultz2023causal,wu2025doflow,xia2025causal}, a world-modeling task that models a system's responses to prescribed inputs rather than extrapolating an observed trajectory.

We call covariates encoding actions or exposures expected to causally affect the target \emph{treatment covariates}, and all others \emph{non-treatment covariates}. The term ``covariate'' denotes a model input and does not imply causal exogeneity~\citep{pearl2009causality}.
Let $\mathbf{Z}_t\in\R^{D_Z}$ and $\mathbf{X}_t\in\R^{D_X}$ collect the treatment and non-treatment covariates, and let $\mc{H}=(Y_{1:T},\mathbf{X}_{1:T},\mathbf{Z}_{1:T})$ denote the observed history.
An interventional forecaster predicts
\begin{equation}
\label{eq:intro-interventional}
    \CondP{Y_{T+1:T+H}}
    {\mc{H},\,\mathrm{do}(\mathbf{Z}_{T+1:T+H}),\,\mathbf{X}_{T+1:T+H}}.
\end{equation}
Current covariate-aware TSFMs instead target the corresponding \emph{observational} conditional, without the $\mathrm{do}$-operator, treating the future treatment path as passively observed; their forecasts therefore need not represent the interventional conditional~\citep{wu2025doflow}.
Nevertheless, such forecasts can be evaluated against intervention-generated outcomes without being interpreted as interventional estimates~\citep{crasson2024training,schultz2023causal,namazi2026prediction}.
Causal forecasters address this mismatch but typically require task-specific training, unlike the zero-shot TSFMs evaluated here~\citep{bica2020crn,melnychuk2022causaltransformer,wu2025doflow}.

A planned path exhibits \emph{covariate novelty} when at least one treatment covariate's future value set is not contained in its historical value set.
We ask: \emph{do covariate-aware TSFMs accurately forecast targets under interventions involving covariate novelty?}
Section~\ref{sec:cov-nov} defines the \scenInDist\ reference and the \scenOOS, \scenSwitch, and \scenOnset\ novelty scenarios.

To our knowledge, no existing benchmark systematically evaluates whether future treatment paths contain values or policies absent from the same series' history.
\CancerTreatment\ is widely used for forecasting under planned treatment sequences~\citep{bica2020crn,melnychuk2022causaltransformer,wu2025doflow}, while \FevBench\ evaluates general covariate-aware forecasting~\citep{shchur2025fev}. Neither benchmark labels evaluation series by covariate novelty.
\appendixref{app:related-work} compares related benchmarks, causal and covariate-aware models, intervention generalization, and distribution shift.

We therefore define covariate novelty by comparing historical and future treatment-covariate value sets for each evaluation series and use it to construct \CovNovBench, a benchmark of $88{,}476$ series from four sources spanning synthetic, semi-synthetic, and real-world data.
The sources emphasize strong treatment effects and controlled or limited hidden confounding to make responses detectable and less ambiguous, without establishing causal identification.
We evaluate six covariate-aware TSFMs zero-shot using point and directional metrics and find that no model is consistently reliable.

\section{Evaluation framework}
\label{sec:evaluation-framework}

\subsection{Covariate novelty}
\label{sec:cov-nov}

Covariate novelty compares historical and future values of each treatment covariate within a single series.
For integers $a\le b$, let $[a:b]=\{a,\ldots,b\}$ and $[b]=[1:b]$.
For each treatment covariate $j\in[D_Z]$ and any nonempty time-index set $\mc{S}$, define
\begin{equation}
  \label{eq:value-set}
  \mc{V}_j(\mc{S})
  =
  \begin{cases}
    \sbr{
    \min_{t\in\mc{S}}Z_{j,t},
    \max_{t\in\mc{S}}Z_{j,t}
    },
    & \text{if numerical},\\[1mm]
    \{Z_{j,t}:t\in\mc{S},\, Z_{j,t}\ne0\},
    & \text{if categorical}.
  \end{cases}
\end{equation}
For categorical covariates, $0$ denotes inactivity and nonzero values denote active policies.
Let $\mc{V}^{\mathrm{hist}}_j=\mc{V}_j([T])$ and $\mc{V}^{\mathrm{fut}}_j=\mc{V}_j([T+1:T+H])$ be the historical and future value sets.
A future treatment path is \scenInDist\ if $\mc{V}^{\mathrm{fut}}_j\subseteq\mc{V}^{\mathrm{hist}}_j$ for every $j\in[D_Z]$, and novel otherwise.
Novelty is assessed per covariate and series, independently of other series or pretraining data.

\paragraph{Numerical treatment covariates.}
A future numerical range extending below the historical minimum or above the historical maximum is \scenOOS.
This setting tests forecasting under numerical treatment extrapolation.
For example, with historical treatment doses 1 and 3, a future dose of 2 stays within range and 4 is Out-of-Support.

\paragraph{Categorical treatment covariates.}
An unseen active policy gives \scenOnset\ after an inactive history ($\mc{V}^{\mathrm{hist}}_j=\emptyset$ and $\mc{V}^{\mathrm{fut}}_j\ne\emptyset$) or \scenSwitch\ after an active history ($\mc{V}^{\mathrm{hist}}_j\ne\emptyset$ and $\mc{V}^{\mathrm{fut}}_j\not\subseteq\mc{V}^{\mathrm{hist}}_j$).
Both cases test forecasting under a supplied unseen policy. Switch histories include active treatments that may inform the forecast.

Appendix~\ref{app:anecdotes} grounds this taxonomy with concrete ``what-if'' queries and representative \CancerTreatment\ trajectories.

\subsection{Point and directional forecast accuracy}
\label{sec:metrics}

We report RMSSE~\citep{makridakis2022m5} for scaled point error, with $1$ marking the stabilized historical one-step error scale.
DA measures sign agreement between predicted and realized target changes over the horizon, which matters when decisions depend on whether the target will increase or decrease~\citep{namazi2026prediction}.
CDA restricts DA to steps with a categorical policy change or a reversal in numerical treatment direction.
Lower RMSSE and higher DA and CDA are better, with $0.5$ as the random-direction reference for DA and CDA.
See \appendixref{app:metric-details} for exact definitions.

\section{\CovNovBench}
\label{sec:benchmark}

Table~\ref{tab:datasets} summarizes supported dataset--scenario pairs.
Where possible, ID and novel scenarios share the same historical treatment settings to ensure fair comparisons, as detailed in \appendixref{app:dataset-registries}.

For categorical datasets, we operationalize the policy-level definitions using one binary indicator per treatment type.
A future policy is novel when it activates a treatment type absent from the history.
The registry labels this transition \scenOnset\ if no treatment type was active in the history and \scenSwitch\ if at least one was active.

\begin{table}[h!]
\centering
\caption{
  \CovNovBench datasets, totaling $88{,}476$ evaluation series.
  tr./st.\ denote trending/stable targets, and c/n denote categorical/numerical treatment covariates.
  ID, OOS, Sw, and On denote the four scenarios; observational datasets provide weaker causal guarantees.
}
\label{tab:datasets}
\footnotesize
\setlength{\tabcolsep}{2pt}
\begin{tabularx}{\linewidth}{@{}llYYcccccr@{}}
\toprule
& & & & \multicolumn{4}{c}{Scenarios} & & \\
\cmidrule(lr){5-8}
Dataset & Type & Target (dyn.) & Treatment cov.\ (type)
  & ID & OOS & Sw & On & $T/H$ & \#series \\
\midrule
\CancerTreatment & Synth.\ & Log tumor volume (tr.) & Radio/chemo dose (c+n)
  & \cmark & \cmark & \cmark & \cmark & 50/10 & 12{,}548 \\
\CausalRivers & Real & Discharge (tr.) & Upstream discharge (n)
  & \cmark & \cmark & \xmark & \xmark & 100/20 & 38{,}992 \\
\MIMICReal & Real & Blood pressure (st.) & Vasopressor/vent.\ (c)
  & \cmark & \xmark & \cmark & \cmark & 30/10 & 7{,}936 \\
\MIMICSynth & Semi-synth.\ & Synth.\ outcome (st.) & Synth.\ policy (c)
  & \cmark & \xmark & \cmark & \cmark & 50/10 & 29{,}000 \\
\bottomrule
\end{tabularx}
\end{table}

\paragraph{\CancerTreatment.}
We adapt the PK/PD tumor-growth simulator of \citet{geng2017prediction} to forecast log tumor volume.
Its treatment policy is no treatment, radiotherapy, chemotherapy, or both, with numerical dose covariates for active treatments.
By varying both policy and dose paths, the simulator supports all four benchmark scenarios.

\paragraph{\CausalRivers.}
\CausalRivers~\citep{stein2025causalrivers} links $666$ German gauging stations in a topology-based causal graph.
We forecast downstream discharge from lagged upstream discharges and static inputs such as distance between gauges.
Small departures from historical upstream ranges remain in ID. Larger departures in either direction define moderate and extreme \scenOOS\ (\appendixref{app:causalrivers}).

\paragraph{\MIMICReal and \MIMICSynth.}
Both use MIMIC-III~\citep{johnson2016mimiciii} through MIMIC-Extract~\citep{wang2020mimic}, following \citet{melnychuk2022causaltransformer}.
\MIMICReal forecasts diastolic blood pressure from the vasopressor/ventilation policy; its treatment associations are suggestive, not causally identified.
\MIMICSynth pairs real vitals with simulated treatments and one simulated outcome; both datasets use categorical treatment policies, so \scenOOS is inapplicable.

\section{Empirical evaluation}
\label{sec:experiments}

We evaluate six covariate-aware TSFMs zero-shot (three tabular and three non-tabular, Appendices~\ref{app:exp-setup} and~\ref{app:access-reproducibility}).
Full dataset--scenario metric tables (RMSSE, DA, and CDA) are reported in Appendix ~\ref{app:full-results}).
Figure~\ref{fig:results-main}(a) gives a benchmark-wide summary.

\begin{figure}[t]
\centering
\includegraphics[width=0.92\linewidth]{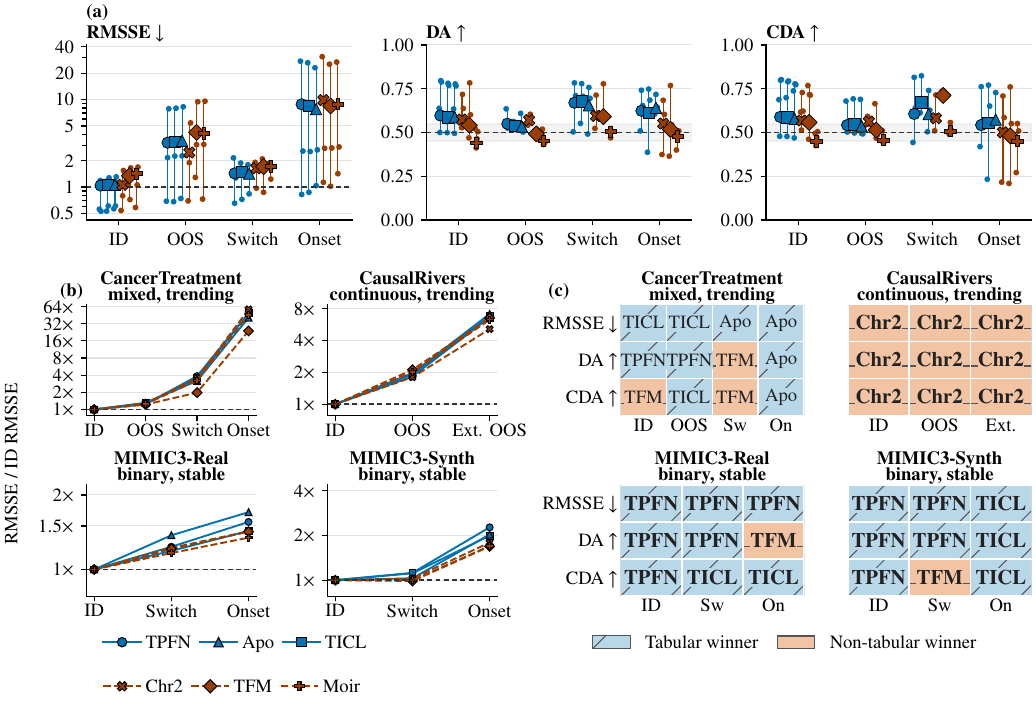}
\caption{
Main results.
  (a) Absolute metrics. Large markers show means weighted by series count.
Small markers show dataset--scenario means, with stems spanning their range.
For Onset, each model has three small markers, one each for \CancerTreatment, \MIMICReal, and \MIMICSynth.
CDA means use series with nonempty transition masks.
(b) RMSSE relative to the same-model, same-dataset \scenInDist\ result.
(c) Best observed model per dataset--scenario--metric cell; color and hatch indicate model family.
Abbreviations: TPFN=\TabPFNvTS, TICL=\TabICL, Apo=\ApolloPFN, Chr2=\ChronosCov, TFM=\TimesFMCov, and Moir=\MoiraiCov.
}
\label{fig:results-main}
\end{figure}

\paragraph{Are TSFMs accurate under covariate novelty?}

No evaluated model is consistently accurate.
All weighted RMSSE averages exceed \(1\), though some dataset--scenario means fall below \(1\).
Many DA and CDA means are near \(0.5\) (Figure~\ref{fig:results-main}(a)).

RMSSE was higher than the \scenInDist\ value in \(52\) of \(54\) comparisons within the same model and dataset (Figure~\ref{fig:results-main}(b)).
The two exceptions are \TimesFMCov\ and \MoiraiCov\ on \MIMICSynth\ \scenSwitch, at or slightly below their \scenInDist\ values.
The RMSSE ratios relative to \scenInDist\ are smaller on \MIMICReal, reach roughly \(24\)--\(57\times\) on \CancerTreatment\ \scenOnset, and increase from moderate to extreme \scenOOS\ for every \CausalRivers\ model.

\paragraph{Do winners vary with dataset characteristics?}

Model winners vary by dataset, and neither family has the best observed mean across all dataset types (Figure~\ref{fig:results-main}(c)).
\ChronosCov\ has the best observed mean in all nine numerical-covariate, trending-target \CausalRivers\ cells; tabular models do so in \(16\) of \(18\) binary-covariate, stable-target MIMIC cells, except for \TimesFMCov\ on \MIMICReal\ \scenOnset\ DA and \MIMICSynth\ \scenSwitch\ CDA.
On \CancerTreatment\ (categorical and numerical treatments; trending target), tabular models have the lowest observed RMSSE mean in all four scenarios, while directional winners vary.
These are descriptive associations because dataset characteristics vary jointly.

\section{Conclusion}
\label{sec:discussion}

We introduced \CovNovBench, a benchmark of $88{,}476$ series drawn from both synthetic and real-world sources for evaluating covariate-aware TSFMs when future treatments contain values or policies absent from the history.
Our zero-shot evaluation of six TSFMs shows that point error increases under novelty in nearly all matched comparisons, directional accuracy frequently approaches the random-direction reference, and no model or model family is consistently best across datasets, leaving substantial room for improvement in forecasting under novel interventions.
Promising directions include retrieval-augmented few-shot forecasting over series where the novel treatment was observed, and conditioning on textual descriptions of each treatment and its expected effect.

\bibliographystyle{plainnat}
\bibliography{citations}

\appendix

\FloatBarrier
\section{Full numerical results}
\label{app:full-results}

Tables~\ref{tab:full-rmsse}, \ref{tab:full-da}, and~\ref{tab:full-cda} list the scores underlying Figure~\ref{fig:results-main}.
For \CausalRivers, the unmarked out-of-support row is moderate and ``Ext.'' denotes extreme.
Appendix~\ref{app:baseline-aggregation} describes aggregation.

\begin{table}[htbp]
\centering
\caption{Full RMSSE ($\downarrow$) results across scenarios and datasets.
Best per benchmark-cell row is in \textbf{bold}; \emph{Average} rows are sample-count weighted over the displayed cells in each scenario block.}
\label{tab:full-rmsse}
\footnotesize
\setlength{\tabcolsep}{1.4pt}
\begin{tabular}{@{}llrrrrrr@{}}
\toprule
Scenario & Dataset & \TabPFNvTS & \TabICL & \TimesFMCov & \MoiraiCov & \ChronosCov & \ApolloPFN \\
\midrule
\multirow{5}{*}{\scenInDist}
 & \CancerTreatment   & 0.56 & \textbf{0.53} & 1.06 & 0.59 & 0.54 & 0.57 \\
 & \CausalRivers      & 1.19 & 1.18 & 1.44 & 1.53 & \textbf{1.06} & 1.16 \\
 & \MIMICReal         & \textbf{0.53} & 0.61 & 0.72 & 1.06 & 0.80 & 0.61 \\
 & \MIMICSynth        & \textbf{1.14} & 1.28 & 1.66 & 1.69 & 1.55 & 1.31 \\
\cmidrule(l){2-8}
 & \textit{Average}   & \textbf{1.04} & 1.07 & 1.35 & 1.42 & 1.06 & 1.06 \\
\midrule
\multirow{4}{*}{\scenOOS}
 & \CancerTreatment   & 0.69 & \textbf{0.68} & 1.29 & 0.73 & 0.70 & 0.74 \\
 & \CausalRivers      & 2.17 & 2.26 & 3.05 & 3.07 & \textbf{1.90} & 2.27 \\
 & \CausalRivers (Ext.) & 7.85 & 7.97 & 9.44 & 9.56 & \textbf{5.41} & 8.25 \\
\cmidrule(l){2-8}
 & \textit{Average}   & 3.21 & 3.29 & 4.21 & 4.13 & \textbf{2.48} & 3.38 \\
\midrule
\multirow{4}{*}{\scenSwitch}
 & \CancerTreatment   & 2.16 & 1.88 & 2.10 & 1.89 & 1.92 & \textbf{1.80} \\
 & \MIMICReal         & \textbf{0.65} & 0.72 & 0.87 & 1.23 & 0.97 & 0.84 \\
 & \MIMICSynth        & \textbf{1.27} & 1.42 & 1.64 & 1.69 & 1.59 & 1.35 \\
\cmidrule(l){2-8}
 & \textit{Average}   & 1.44 & 1.50 & 1.71 & 1.72 & 1.64 & \textbf{1.43} \\
\midrule
\multirow{4}{*}{\scenOnset}
 & \CancerTreatment   & 27.39 & 26.28 & 25.31 & 26.76 & 30.83 & \textbf{23.05} \\
 & \MIMICReal         & \textbf{0.82} & 0.87 & 1.02 & 1.42 & 1.13 & 1.04 \\
 & \MIMICSynth        & 2.58 & \textbf{2.55} & 2.80 & 2.88 & 2.77 & 2.66 \\
\cmidrule(l){2-8}
 & \textit{Average}   & 8.81 & 8.51 & 8.44 & 8.89 & 9.85 & \textbf{7.77} \\
\bottomrule
\end{tabular}
\end{table}

\begin{table}[htbp]
\centering
\caption{Full Directional Accuracy (DA, $\uparrow$) results across scenarios and datasets.
Best per benchmark-cell row is in \textbf{bold}; \emph{Average} rows are sample-count weighted over the displayed cells in each scenario block.}
\label{tab:full-da}
\footnotesize
\setlength{\tabcolsep}{1.4pt}
\begin{tabular}{@{}llrrrrrr@{}}
\toprule
Scenario & Dataset & \TabPFNvTS & \TabICL & \TimesFMCov & \MoiraiCov & \ChronosCov & \ApolloPFN \\
\midrule
\multirow{5}{*}{\scenInDist}
 & \CancerTreatment   & \textbf{.638} & .563 & .568 & .520 & .559 & .634 \\
 & \CausalRivers      & .501 & .499 & .470 & .412 & \textbf{.528} & .496 \\
 & \MIMICReal         & \textbf{.796} & .781 & .783 & .464 & .718 & .777 \\
 & \MIMICSynth        & \textbf{.789} & .769 & .601 & .505 & .637 & .766 \\
\cmidrule(l){2-8}
 & \textit{Average}   & \textbf{.597} & .586 & .542 & .441 & .574 & .587 \\
\midrule
\multirow{4}{*}{\scenOOS}
 & \CancerTreatment   & \textbf{.635} & .563 & .508 & .519 & .542 & .609 \\
 & \CausalRivers      & .526 & .521 & .478 & .431 & \textbf{.567} & .519 \\
 & \CausalRivers (Ext.) & .529 & .556 & .510 & .441 & \textbf{.604} & .503 \\
\cmidrule(l){2-8}
 & \textit{Average}   & .550 & .539 & .492 & .453 & \textbf{.570} & .535 \\
\midrule
\multirow{4}{*}{\scenSwitch}
 & \CancerTreatment   & .504 & .552 & \textbf{.581} & .505 & .523 & .491 \\
 & \MIMICReal         & \textbf{.783} & .779 & .778 & .470 & .713 & .757 \\
 & \MIMICSynth        & \textbf{.710} & .706 & .587 & .506 & .609 & .695 \\
\cmidrule(l){2-8}
 & \textit{Average}   & .670 & \textbf{.677} & .591 & .505 & .595 & .655 \\
\midrule
\multirow{4}{*}{\scenOnset}
 & \CancerTreatment   & .509 & .387 & .365 & .400 & .374 & \textbf{.605} \\
 & \MIMICReal         & .741 & .748 & \textbf{.768} & .470 & .684 & .718 \\
 & \MIMICSynth        & .653 & \textbf{.685} & .545 & .509 & .603 & .645 \\
\cmidrule(l){2-8}
 & \textit{Average}   & .624 & .614 & .518 & .478 & .551 & \textbf{.641} \\
\bottomrule
\end{tabular}
\end{table}

\begin{table}[htbp]
\centering
\caption{Full Conditional Directional Accuracy (CDA, $\uparrow$) results across scenarios and datasets.
Best per benchmark-cell row is in \textbf{bold}; \emph{Average} rows are sample-count weighted over the displayed cells in each scenario block.}
\label{tab:full-cda}
\footnotesize
\setlength{\tabcolsep}{1.4pt}
\begin{tabular}{@{}llrrrrrr@{}}
\toprule
Scenario & Dataset & \TabPFNvTS & \TabICL & \TimesFMCov & \MoiraiCov & \ChronosCov & \ApolloPFN \\
\midrule
\multirow{5}{*}{\scenInDist}
 & \CancerTreatment   & .688 & .699 & \textbf{.768} & .495 & .617 & .737 \\
 & \CausalRivers      & .479 & .478 & .462 & .425 & \textbf{.514} & .469 \\
 & \MIMICReal         & \textbf{.798} & .789 & .717 & .474 & .728 & .778 \\
 & \MIMICSynth        & \textbf{.802} & .794 & .713 & .505 & .622 & .773 \\
\cmidrule(l){2-8}
 & \textit{Average}   & \textbf{.589} & .586 & .558 & .449 & .569 & .579 \\
\midrule
\multirow{4}{*}{\scenOOS}
 & \CancerTreatment   & .680 & \textbf{.692} & .665 & .510 & .592 & .689 \\
 & \CausalRivers      & .503 & .497 & .468 & .439 & \textbf{.547} & .495 \\
 & \CausalRivers (Ext.) & .508 & .525 & .485 & .451 & \textbf{.570} & .492 \\
\cmidrule(l){2-8}
 & \textit{Average}   & .542 & .546 & .515 & .457 & \textbf{.562} & .536 \\
\midrule
\multirow{4}{*}{\scenSwitch}
 & \CancerTreatment   & .443 & .612 & \textbf{.736} & .490 & .510 & .504 \\
 & \MIMICReal         & .815 & \textbf{.824} & .734 & .557 & .714 & .740 \\
 & \MIMICSynth        & .642 & .683 & \textbf{.704} & .511 & .596 & .628 \\
\cmidrule(l){2-8}
 & \textit{Average}   & .606 & .673 & \textbf{.712} & .508 & .581 & .605 \\
\midrule
\multirow{4}{*}{\scenOnset}
 & \CancerTreatment   & .420 & .233 & .209 & .271 & .217 & \textbf{.571} \\
 & \MIMICReal         & .762 & \textbf{.771} & .760 & .551 & .708 & .728 \\
 & \MIMICSynth        & .562 & \textbf{.651} & .551 & .509 & .585 & .555 \\
\cmidrule(l){2-8}
 & \textit{Average}   & .543 & .554 & .481 & .451 & .501 & \textbf{.574} \\
\bottomrule
\end{tabular}
\end{table}

\FloatBarrier

\section{Additional related work}
\label{app:related-work}

This appendix expands the compact positioning discussion in the main paper.

\paragraph{Forecasting benchmarks.}
\Mthree~\citep{makridakis2000m3} and \Mfour~\citep{makridakis2020m4} mainly forecast targets from their own histories, while \Mfive~\citep{makridakis2022m5} adds a large retail panel with price and calendar inputs; the \MonashArchive~\citep{godahewa2021monash} and \GIFTEval~\citep{aksu2024gifteval} broaden domain, frequency, and scale coverage.
General benchmarks do not define treatment covariates, while resources with known future inputs, including \Mfive, electricity-price forecasting with day-ahead load and generation forecasts~\citep{lago2021electricity}, and \FevBench's covariate-aware tasks~\citep{shchur2025fev}, treat those inputs as observational continuations rather than distinguish passively observed from actively set paths.
None annotates whether a future treatment path is novel relative to its own series history, the axis targeted by \CovNovBench.

\paragraph{Covariate-aware foundation models.}
Early TSFMs such as \ChronosUni~\citep{ansari2024chronos} and \TimesFMUni~\citep{das2024timesfm} mainly studied univariate zero-shot forecasting; \ChronosCov~\citep{ansari2025chronos}, \TimesFMCov~\citep{das2024timesfm}, and \MoiraiCov~\citep{woo2024unified} accept covariates, as do tabular models adapted by representing forecast steps as supervised examples: \ApolloPFN~\citep{potapczynski2026time}, \TabPFNvTS~\citep{hoo2025tables}, and \TabICL~\citep{qu2026tabiclv2}.
These interfaces make natural baselines, but their observational forecasts need not equal the interventional conditional under actively set treatment paths; we therefore evaluate them as frozen observational baselines, not causal forecasters.

\paragraph{Causal forecasting and causal foundation models.}
Models for specified treatment regimes include task-specific \RMSN~\citep{lim2018rmsn}, \CRN~\citep{bica2020crn}, \CT~\citep{melnychuk2022causaltransformer}, and \IGCNet~\citep{hess2026igcnet}, as well as \DoFlow~\citep{wu2025doflow} and \CaTSG~\citep{xia2025causal} for interventional or counterfactual trajectory distributions; these approaches require task-specific training, causal assumptions, or structural information.
Across tasks, \DoPFN~\citep{robertson2025dopfn}, \CausalPFN~\citep{balazadeh2025causalpfn}, and \CausalFM~\citep{ma2025causalfm} apply foundation-model or in-context ideas mainly to tabular effect estimation, while \CausalTimePrior~\citep{thumm2026causaltimeprior} provides a temporal causal prior and a proof-of-concept PFN trained on synthetic temporal structural causal models; none provides a general covariate-aware TSFM for zero-shot multi-step interventional forecasting.
Theory distinguishes statistical from causal forecast risk~\citep{vankadara2022causal}, and empirical work studies changed pricing policies, actions outside the observational distribution, and paired factual and counterfactual insulin plans~\citep{schultz2023causal,crasson2024training,namazi2026prediction}.
These works motivate intervention evaluation but provide neither our future-path taxonomy nor a cross-dataset benchmark of frozen covariate-aware TSFMs; \CovNovBench\ instead evaluates supplied novel paths against simulator-generated or observationally filtered outcomes, depending on the dataset.

\paragraph{Novel interventions.}
Zero-shot causal learning also studies unseen interventions: \CaML~\citep{nilforoshan2023zero} uses semantic and structured intervention representations for personalized effects and combinations, while compositional sequential-intervention methods exploit structure to handle sparse or unseen sequences~\citep{yu2024structured}.
These settings resemble \scenOnset\ and \scenSwitch, but they develop causal or structured intervention models, whereas \CovNovBench\ evaluates whether existing frozen forecasters can use a supplied novel treatment path.

\paragraph{Distribution shift and extrapolation.}
Classical covariate shift assumes a fixed conditional under input-distribution change~\citep{shimodaira2000improving}; time-series OOD work studies broader shifts across time, domains, and generating regimes~\citep{wu2026out}; and regression or intervention extrapolation addresses prediction beyond observed support~\citep{shen2025engression,saengkyongam2024identifying}.
These notions define shift relative to a population, dataset, domain, or action distribution, whereas our label compares a supplied horizon with the same series' history, separating treatment-path novelty from broader target non-stationarity or dataset shift.

\section{Metric details}
\label{app:metric-details}

\paragraph{RMSSE.}
For one evaluation series with target history $Y_{1:T}$ and horizon targets $Y_{T+1:T+H}$, $\hat{Y}_{T+h}$ is the median predicted quantile and $\eps=1\times10^{-2}$ stabilizes denominators in all three metrics.
RMSSE~\citep{makridakis2022m5} scales forecast mean squared error by historical one-step variation,
\begin{equation}
\label{eq:rmsse}
\mathrm{RMSSE}
=
\sqrt{
\frac{
\frac{1}{H}\sum_{h=1}^{H}
\p{\hat{Y}_{T+h}-Y_{T+h}}^2
}{
\frac{1}{T-1}\sum_{t=2}^{T}
\p{Y_t-Y_{t-1}}^2+\eps
}
}.
\end{equation}
The denominator is the mean squared historical one-step error plus $\eps$. Lower RMSSE is better, and values above $1$ indicate that the forecast mean squared error exceeds this stabilized reference.

\paragraph{DA and CDA.}
DA measures directional agreement, which matters when an incorrect direction can misstate whether an outcome improves or worsens~\citep{namazi2026prediction}.
Set $\hat{Y}_T=Y_T$, and define $r_h=(Y_{T+h}-Y_{T+h-1})/(\abs{Y_{T+h-1}}+\eps)$ and $\hat r_h$ analogously.
Let $\mc{M}\subseteq[H]$ be the union of treatment-covariate transition steps---categorical policy changes or numerical direction reversals---counting a step once even if several covariates transition.
DA averages sign agreement over the horizon, whereas CDA restricts it to $\mc{M}$:
\begin{equation}
\label{eq:da}
\mathrm{DA}
=
\frac{1}{H}\sum_{h=1}^{H}
\mathbf{1}\!\sbr{
\operatorname{sign}(\hat{r}_h)
=
\operatorname{sign}(r_h)
},
\qquad
\mathrm{CDA}
=
\frac{1}{\abs{\mc{M}}}
\sum_{h\in\mc{M}}
\mathbf{1}\!\sbr{
\operatorname{sign}(\hat{r}_h)
=
\operatorname{sign}(r_h)
}.
\end{equation}
Both lie in $[0,1]$ and higher is better; the mask controls only where CDA is evaluated and is distinct from the novelty criterion.

\paragraph{Aggregation.}
All three metrics are computed per evaluation series.
RMSSE and DA are averaged arithmetically over all series assigned to each dataset--scenario result.
CDA is computed only when $\mc{M}\ne\emptyset$; series with no designated treatment transition are excluded from the dataset--scenario CDA mean.

\section{Benchmark interpretation, construction, and scenario-subset registries}
\label{app:anecdotes}
\label{app:dataset-registries}

\begin{wrapfigure}{r}{0.5\textwidth}
    \centering
    \includegraphics[width=0.5\textwidth]{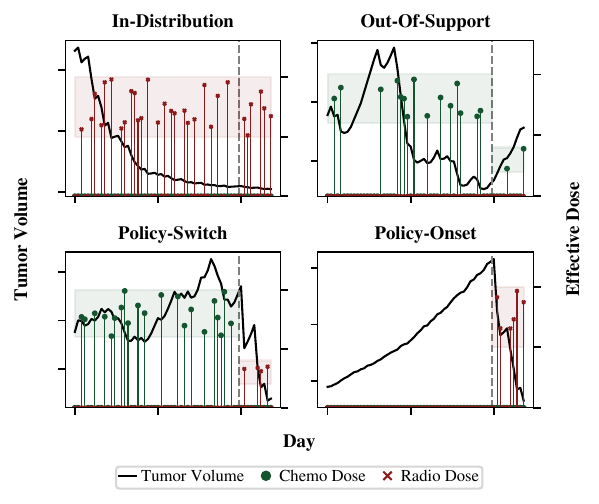}
    \caption{Representative \CancerTreatment\ trajectories for Table~\ref{tab:intro-covnov-taxonomy}. Black curves show tumor volume, green and red stems show chemotherapy and radiotherapy doses, dashed lines mark the forecast boundary, and shading marks registered active-dose ranges.}
    \label{fig:intro-covnov-scenarios}
\end{wrapfigure}

Table~\ref{tab:intro-covnov-taxonomy} translates the taxonomy into concrete supply-chain and \CancerTreatment\ \emph{what-if} questions, and Figure~\ref{fig:intro-covnov-scenarios} visualizes the \CancerTreatment\ cases.

\begin{table}[htbp]
\centering
\caption{Illustrative questions for the non-novel \scenInDist\ reference and three novelty scenarios, ordered nominally from easy to hard. BOGO denotes a buy-one-get-one offer; the \CancerTreatment\ cases appear in Figure~\ref{fig:intro-covnov-scenarios}.}
\label{tab:intro-covnov-taxonomy}
\footnotesize
\setlength{\tabcolsep}{1mm}
\begin{tabularx}{\linewidth}{@{}p{0.155\linewidth}>{\RaggedRight\arraybackslash}p{0.20\linewidth}YY@{}}
\toprule
  & & \multicolumn{2}{c}{Query: \emph{What if we \ldots?}} \\
\cmidrule(lr){3-4}
\makecell[l]{Scenario\\(easy $\downarrow$ hard)}
  & Scenario pattern
  & \makecell[l]{Supply chain\\(demand)}
  & \makecell[l]{\CancerTreatment\\(tumor volume)} \\
\midrule
\scenInDist
  & Future values seen in history
  & Repeat a previously used discount?
  & Continue a prior radiotherapy dose? \\
\addlinespace[2pt]
\scenOOS
  & Future range outside history
  & Offer $40\%$ after only $<20\%$ discounts?
  & Use a chemo dose outside history? \\
\addlinespace[2pt]
\scenSwitch
  & Active history $\to$ new policy
  & Replace a discount with BOGO?
  & Switch from chemo to radiotherapy? \\
\addlinespace[2pt]
\scenOnset
  & Inactive $\to$ active policy
  & Launch the product's first discount?
  & Start radiotherapy for a new patient? \\
\bottomrule
\end{tabularx}
\end{table}

Within a dataset--scenario pair, an \emph{evaluation subset} fixes the treatment covariates and their exact history-to-horizon pattern.
Each dataset's scenario-subset registry is fixed before generation or observational filtering; registered subsets are pooled for dataset--scenario reporting, except that moderate and extreme \CausalRivers\ \scenOOS\ remain separate results.
Categorical registries exclude inactive-to-inactive horizons and de-escalation-only transitions because neither introduces a novel active value.

\paragraph{History comparability.}
In \CancerTreatment, \scenOOS\ and \scenSwitch\ subsets have \scenInDist\ counterparts with the same allowed historical treatment types and dose ranges (Table~\ref{tab:cancer-subsets}).
For example, radiotherapy histories in \scenInDist\ and \scenOOS\ both use doses in $[1,2]$.
The \scenSwitch\ subsets in \MIMICReal\ and \MIMICSynth\ share historical active treatment sets with their \scenInDist\ counterparts (Tables~\ref{tab:mimic-real-subsets} and~\ref{tab:mimic-synth-subsets}).
\CausalRivers\ uses the same graph types and windowing procedure for \scenInDist\ and \scenOOS\ (Table~\ref{tab:causalrivers-subsets}).
\scenOnset\ starts from untreated histories and has no corresponding \scenInDist\ subset.
The reported scenario means pool separately generated or selected windows.

\subsection{\CancerTreatment}
\label{app:cancer}

\paragraph{Construction and assumptions.}
This synthetic dataset uses the PK/PD tumor-growth model of \citet{geng2017prediction}, adapted by \citet{bica2020crn,melnychuk2022causaltransformer}, to forecast log tumor volume from binary radiotherapy/chemotherapy actions, continuous doses, and a static patient type.
Radiotherapy acts immediately, chemotherapy decays over time, and treatment assignment depends on recent tumor diameter, creating controlled time-varying confounding; the specified-policy outcomes are simulator semantics rather than clinical claims, and very strong treatment can approach the simulator's boundary near zero tumor volume.
For each registry row, split-window controls constrain history and horizon treatment types and doses, disallowed assignment probabilities are zeroed, and $1{,}000$ patient-specific trajectories of $L=60$ daily steps are sampled.
After removing incomplete or infeasible trajectories and shifting treatments by one step, the benchmark uses $T=50$, $H=10$, and retains $12{,}548$ series: $2{,}969$ \scenInDist, $2{,}868$ \scenOOS, $2{,}749$ \scenOnset, and $3{,}962$ \scenSwitch.

\paragraph{Registry.}
Table~\ref{tab:cancer-subsets} lists all $13$ subsets with their history and horizon policies.
In-range radiotherapy and chemotherapy doses are $[1,2]$ and $[3,5]$, while their \scenOOS\ ranges are $[0,1]\cup[2,3]$ and $[0,3]\cup[5,6]$, respectively.

\begin{table}[htbp]
\centering
\caption{\CancerTreatment\ evaluation-subset registry. Radio and chemo denote radiotherapy and chemotherapy.}
\label{tab:cancer-subsets}
\footnotesize
\setlength{\tabcolsep}{4pt}
\begin{tabularx}{\linewidth}{@{}lllY@{}}
\toprule
Scenario & Subset pattern & History policy & Horizon policy \\
\midrule
\scenInDist & radio & radio, dose $[1,2]$ & radio, dose $[1,2]$ \\
\scenInDist & chemo & chemo, dose $[3,5]$ & chemo, dose $[3,5]$ \\
\scenInDist & both & radio + chemo, in-range doses & radio + chemo, in-range doses \\
\midrule
\scenOOS & radio dose & radio, dose $[1,2]$ & radio, dose $[0,1]\cup[2,3]$ \\
\scenOOS & chemo dose & chemo, dose $[3,5]$ & chemo, dose $[0,3]\cup[5,6]$ \\
\scenOOS & both doses & radio + chemo, in-range doses & radio + chemo, out-of-range doses \\
\midrule
\scenOnset & to radio & no treatment & radio, dose $[1,2]$ \\
\scenOnset & to chemo & no treatment & chemo, dose $[3,5]$ \\
\scenOnset & to both & no treatment & radio + chemo, in-range doses \\
\midrule
\scenSwitch & radio to chemo & radio, dose $[1,2]$ & chemo, dose $[3,5]$ \\
\scenSwitch & chemo to radio & chemo, dose $[3,5]$ & radio, dose $[1,2]$ \\
\scenSwitch & radio to both & radio, dose $[1,2]$ & radio + chemo, in-range doses \\
\scenSwitch & chemo to both & chemo, dose $[3,5]$ & radio + chemo, in-range doses \\
\bottomrule
\end{tabularx}
\end{table}

\FloatBarrier

\subsection{\CausalRivers}
\label{app:causalrivers}

\paragraph{Construction and assumptions.}
We use RiversEastGermany from \CausalRivers~\citep{stein2025causalrivers}, containing $15$-minute discharge from $666$ gauges and $651$ directed edges during $2019$--$2023$.
For each pair, chain, or confluence subgraph, the downstream discharge is the target, lagged upstream discharges are treatment covariates, and distance, elevation difference, and same-river status are static inputs.
The topology gives a physically motivated direction but not an identified intervention effect: unobserved rainfall, varying transport lags, and omitted tributaries can confound these observational associations.
We retain gauges with more than $95\%$ valid positive observations, aggregate to $6$-hour means, restrict chain/confluence targets to modeled direct parents, and form windows with $T=100$, $H=20$, and stride $H$.
After removing invalid or near-constant histories, capping windows per subgraph, and shifting upstream covariates by one step, $38{,}992$ series remain: $28{,}532$ \scenInDist, $7{,}258$ moderate \scenOOS, and $3{,}202$ extreme \scenOOS.

\paragraph{Registry.}
For upstream covariates $\mathbf{Z}$, the deviation score is the maximum normalized departure from each covariate's own history range:
\begin{equation}
d(\mathbf{Z})
=
\max_j
\frac{
\max\!\left\{0, Z_{j,\max}^{\mathrm{fut}}-Z_{j,\max}^{\mathrm{hist}}, Z_{j,\min}^{\mathrm{hist}}-Z_{j,\min}^{\mathrm{fut}}\right\}
}{
Z_{j,\max}^{\mathrm{hist}}-Z_{j,\min}^{\mathrm{hist}}
}.
\label{eq:river-deviation}
\end{equation}
Every structure in Table~\ref{tab:causalrivers-subsets} is crossed with $d<0.1$ (\scenInDist), $0.1\le d<1$ (moderate \scenOOS), and $d\ge1$ (extreme \scenOOS); thus small positive departures remain in the reference group.
Continuous discharge has no natural inactive state, so \scenOnset\ and categorical \scenSwitch\ are undefined.

\begin{table}[htbp]
\centering
\caption{\CausalRivers\ registry: three structures crossed with ID, moderate-OOS, and extreme-OOS bands (nine subsets).}
\label{tab:causalrivers-subsets}
\footnotesize
\setlength{\tabcolsep}{5pt}
\begin{tabularx}{\linewidth}{@{}llY@{}}
\toprule
Structure & Treatment covariates & Role \\
\midrule
Pair ($A\!\to\!B$) & one upstream discharge & direct upstream influence \\
Chain ($A\!\to\!B\!\to\!C$) & grandparent and parent discharges & mediated upstream influence \\
Confluence ($A\!\to\!C\!\leftarrow\!B$) & two upstream tributary discharges & simultaneous upstream influence \\
\bottomrule
\end{tabularx}
\end{table}

\subsection{\MIMICReal}
\label{app:mimic-real}

\paragraph{Construction and assumptions.}
We use the hourly MIMIC-Extract~\citep{wang2020mimic} representation of deidentified MIMIC-III~\citep{johnson2016mimiciii}, following \citet{melnychuk2022causaltransformer}.
The target is diastolic blood pressure; vasopressor administration and mechanical ventilation are binary treatment covariates, with $25$ hourly clinical variables and static gender, ethnicity, and age as other inputs after the source imputation and normalization.
Because treatments respond to patient state and counterfactual outcomes are unavailable, confounding by indication prevents interpreting these observational associations as intervention effects.
Eligible ICU stays have at least $40$ hourly observations and are capped at $250$ hours.
\scenOnset\ and \scenSwitch\ boundaries use the first activation of a horizon-novel treatment type, with the preceding history checked against the registry; \scenInDist\ uses a persistent active policy, and one stay may supply distinct qualifying transitions.
After a one-step treatment shift, the $T=30$, $H=10$ windows total $7{,}936$: $6{,}487$ \scenInDist, $934$ \scenOnset, and $515$ \scenSwitch.

\paragraph{Registry.}
The two indicators define none, vasopressor-only, ventilation-only, and joint policies.
A vasopressor-to-ventilation transition is \scenOnset\ for the ventilation indicator and is registered as \scenSwitch\ for the overall policy because vasopressors were active in the history.
Table~\ref{tab:mimic-real-subsets} preserves all ten registered transitions; all-off horizons and de-escalation-only changes are excluded.

\begin{table}[htbp]
\centering
\caption{\MIMICReal\ scenario-subset registry. Vaso and Vent denote vasopressor and ventilation.}
\label{tab:mimic-real-subsets}
\footnotesize
\setlength{\tabcolsep}{5pt}
\begin{tabularx}{\linewidth}{@{}llY@{}}
\toprule
Scenario & \# subsets & Registered history $\to$ horizon patterns \\
\midrule
\scenInDist & 3 & Vaso$\to$Vaso, Vent$\to$Vent, Both$\to$Both \\
\scenOnset & 3 & None$\to$Vaso, None$\to$Vent, None$\to$Both \\
\scenSwitch & 4 & Vaso$\to$Vent, Vent$\to$Vaso, Vaso$\to$Both, Vent$\to$Both \\
\bottomrule
\end{tabularx}
\end{table}

\subsection{\MIMICSynth}
\label{app:mimic-synth}

\paragraph{Construction and assumptions.}
Following \citet{melnychuk2022causaltransformer}, this semi-synthetic dataset combines $25$ normalized hourly MIMIC-Extract variables and static gender, ethnicity, and age with three simulated binary treatments and one simulated outcome.
The untreated outcome includes smooth temporal structure, covariate dependence, patient heterogeneity, and noise; treatment propensities depend on recent outcomes and covariates, creating controlled time-varying confounding, while effects are additive over finite windows.
The mechanism supports specified-policy outcomes, conditional on the simulator, and reconstruction requires authorized MIMIC-III access.
Eligible stays have at least $60$ hours and are capped at $100$; for each registry row, a history--horizon allow mask constrains assignment and a separately seeded generator samples $1{,}000$ trajectories of $L=60$.
After a one-step treatment shift, the $T=50$, $H=10$ windows total exactly $29{,}000$: $7{,}000$ \scenInDist, $7{,}000$ \scenOnset, and $15{,}000$ \scenSwitch.

\paragraph{Registry.}
For active sets formed from $t_1,t_2,t_3$, Table~\ref{tab:mimic-synth-subsets} compactly specifies all $29$ subsets.

\begin{table}[htbp]
\centering
\caption{\MIMICSynth\ registry ($0$: inactive; distinct $i,j,k$).}
\label{tab:mimic-synth-subsets}
\scriptsize
\setlength{\tabcolsep}{5pt}
\begin{tabularx}{\linewidth}{@{}llY@{}}
\toprule
Scenario & \# subsets & Registered history $\to$ horizon patterns \\
\midrule
\scenInDist & 7 & $S\to S$ for every nonempty $S\subseteq\{t_1,t_2,t_3\}$ \\
\scenOnset & 7 & $0\to S$ for every nonempty $S\subseteq\{t_1,t_2,t_3\}$ \\
\scenSwitch & 15 & six pure replacements $t_i\to t_j$; three single-to-complementary-pair replacements $t_i\to t_jt_k$; three pair-to-complement replacements $t_it_j\to t_k$; three pair-to-triple escalations $t_it_j\to t_1t_2t_3$ \\
\bottomrule
\end{tabularx}
\end{table}

\FloatBarrier

\section{Experimental setup}
\label{app:exp-setup}

\subsection{Models and protocol}
\label{app:model-details}

All six frozen checkpoints are evaluated once in zero-shot mode, independently on each of the $88{,}476$ evaluation series, using its target history, observed covariates, and supplied future covariates; the median probabilistic forecast is the point forecast.
No reference series, benchmark-specific tuning or fine-tuning, supervised adaptation, or test-outcome feedback is used.
Benchmark construction and stochastic inference use fixed seeds; inference uses seed $1$ except for \ChronosCov\ and \MoiraiCov, which use seed $42$, with deterministic backend settings enabled where supported.
The results therefore characterize one fixed benchmark realization rather than repeated-run or alternative-draw uncertainty.
Table~\ref{tab:model-details} consolidates the model-specific settings.

\begin{table}[htbp]
\centering
\caption{Checkpoints, interfaces, and configurations used in zero-shot evaluation.}
\label{tab:model-details}
\footnotesize
\setlength{\tabcolsep}{3pt}
\begin{tabularx}{\linewidth}{@{}>{\RaggedRight\arraybackslash}p{0.15\linewidth}>{\RaggedRight\arraybackslash}p{0.34\linewidth}Y@{}}
\toprule
Model & Checkpoint / interface & Configuration / preprocessing \\
\midrule
\ApolloPFN & Closed, owner-authorized checkpoint; forecast-step tabular examples; MQLoss head outputs nine quantiles. & $12$-layer PFNT, $192$ hidden dimensions, $6$ heads, MLP factor $4$, no bias or dropout; feature grouping $2$, input size $512$, \texttt{norm\_y=False}. The checkpoint is unavailable publicly. \\
\TabPFNvTS & \texttt{tabpfn} with \texttt{tabpfn-v2-regressor.ckpt} (Prior-Labs, HuggingFace); forecast-step tabular examples; nine quantiles from the inverse CDF of a BarDistribution. & $12$-layer PFNT, $192$ hidden dimensions, $6$ heads, no dropout; $5{,}000$ bins from \texttt{borders.npy}, feature grouping $2$, per-series z-normalization (\texttt{norm\_y=True}), per-batch uniform QuantileTransformer, StandardScaler, and rank-$1$ SVD augmentation. NaNs receive indicators; zero-valid-context series receive zero predictions. \\
\TabICL & \texttt{tabicl}; TabICL-v2 checkpoint auto-downloaded from HuggingFace; \texttt{TabICLRegressor.fit()} uses historical covariates and targets, then \texttt{.predict()} uses future covariates. & Eight ensemble estimators; most recent $1{,}024$ history steps retained when needed; no per-series optimization beyond in-context learning. \\
\ChronosCov & \texttt{chronos-forecasting} with \texttt{amazon/chronos-2}; named past/future-covariate dictionaries passed to the native covariate-aware API. & Context determined from target history up to $2{,}048$ tokens; quantiles returned directly. \\
\TimesFMCov & \href{https://github.com/google-research/timesfm}{\texttt{timesfm}} with the public \href{https://huggingface.co/google/timesfm-2.5-200m-pytorch}{TimesFM-2.5 200M checkpoint} through \texttt{TimesFM\_2p5\_200M\_torch}; \texttt{dynamic\_numerical\_covariates} passed to \texttt{forecast\_with\_covariates()}. & Context $1{,}024$, horizon $64$; \texttt{normalize\_inputs}, \texttt{use\_continuous\_quantile\_head}, \texttt{force\_flip\_invariance}, \texttt{infer\_is\_positive}, and \texttt{fix\_quantile\_crossing} set true. Padding is stripped before each call. \\
\MoiraiCov & \href{https://github.com/SalesforceAIResearch/uni2ts}{\texttt{uni2ts}} with \texttt{Salesforce/moirai-1.1-R-large}, loaded by \texttt{MoiraiModule.from\_pretrained} and \texttt{MoiraiForecast}; \texttt{feat\_dynamic\_real} spans context and horizon. & Context $512$, patch size $16$, and $100$ forecast samples converted to quantiles. NaN covariates are zeroed with an observed-mask indicator; seed $42$ makes sampling repeatable. \\
\bottomrule
\end{tabularx}
\end{table}

\paragraph{Implementation.}
All wrappers share a PyTorch Lightning harness driven by a dataset-registry YAML with one combined-Parquet entry per scenario; the runner generates scenario configurations and writes per-sample and aggregate metric CSVs.
LLM-based assistance was limited to code implementation of the author-defined scenario-subset registries and filtering pipeline.

\subsection{Compute resources}
\label{app:compute-resources}

All $78$ model--dataset--scenario evaluations ran on one AWS \href{https://docs.aws.amazon.com/ec2/latest/instancetypes/ac.html}{\texttt{p4d.24xlarge}} instance with $96$ vCPUs, $1{,}152$ GiB memory, eight $40$-GiB NVIDIA A100 GPUs, and eight $1$-TB NVMe SSDs.
Table~\ref{tab:inference-time} reports the recorded elapsed forward-pass time for each evaluation.
The underlying unrounded durations total $13{,}604.2$ seconds ($3.78$ hours); model loading, preprocessing, and broader project compute are excluded.

\begin{table}[htbp]
\centering
\caption{Forward-pass inference times (s); totals sum unrounded entries and are rounded to $0.1$ s.}
\label{tab:inference-time}
\footnotesize
\setlength{\tabcolsep}{2.6pt}
\begin{tabular}{@{}llrrrrrr@{}}
\toprule
Scenario & Dataset & \makecell{\texttt{TabPFN-}\\\texttt{TS}} & \makecell{\texttt{TabICL-}\\\texttt{v2}} & \makecell{\texttt{TimesFM-}\\\texttt{2.5}} & \makecell{\texttt{Moirai-}\\\texttt{1.1}} & \makecell{\texttt{Chronos-}\\\texttt{2}} & \makecell{\texttt{Apollo}\\\texttt{PFN}} \\
\midrule
\multirow{4}{*}{\scenInDist}
 & \CancerTreatment & 23.5 & 805.6 & 98.7 & 30.3 & 20.1 & 17.3 \\
 & \CausalRivers    & 370.7 & 1986.2 & 521.6 & 49.9 & 31.4 & 88.1 \\
 & \MIMICReal       & 31.0 & 584.6 & 192.9 & 74.7 & 23.1 & 18.8 \\
 & \MIMICSynth      & 26.2 & 557.9 & 232.8 & 76.6 & 28.4 & 27.7 \\
\midrule
\multirow{3}{*}{\scenOOS}
 & \CancerTreatment      & 22.9 & 691.2 & 94.3 & 29.6 & 19.5 & 17.4 \\
 & \CausalRivers         & 92.9 & 506.9 & 149.4 & 36.2 & 23.7 & 36.5 \\
 & \CausalRivers\ (Ext.) & 48.7 & 522.0 & 81.9 & 29.1 & 27.0 & 34.3 \\
\midrule
\multirow{3}{*}{\scenSwitch}
 & \CancerTreatment & 22.3 & 627.4 & 94.3 & 29.9 & 21.3 & 18.4 \\
 & \MIMICReal       & 12.0 & 276.2 & 26.0 & 36.3 & 14.8 & 11.4 \\
 & \MIMICSynth      & 35.7 & 1051.2 & 380.1 & 118.9 & 30.5 & 29.0 \\
\midrule
\multirow{3}{*}{\scenOnset}
 & \CancerTreatment & 19.8 & 671.7 & 92.0 & 30.1 & 18.7 & 18.1 \\
 & \MIMICReal       & 14.7 & 511.1 & 35.2 & 54.2 & 16.0 & 11.9 \\
 & \MIMICSynth      & 27.6 & 538.9 & 197.5 & 75.8 & 27.2 & 26.3 \\
\midrule
 & \textit{Total} & 748.3 & 9330.8 & 2196.5 & 671.7 & 301.7 & 355.3 \\
\bottomrule
\end{tabular}
\end{table}

\subsection{Result aggregation}
\label{app:baseline-aggregation}

Each dataset--scenario RMSSE and DA value is the arithmetic mean over all of its per-series values; CDA is averaged over the series with nonempty transition masks.
The benchmark-wide averages in Figure~\ref{fig:results-main} weight each dataset--scenario result by its evaluation-series count; moderate and extreme \CausalRivers\ \scenOOS\ results contribute separately.
All benchmark-wide averages are descriptive; matched same-model, same-dataset comparisons against \scenInDist\ are the primary evidence.

\section{Access and reproducibility boundaries}
\label{app:access-reproducibility}

\paragraph{Data and licenses.}
The MIT-licensed Causal Transformer code supports both generated datasets, and MIMIC-Extract is MIT-licensed; \CausalRivers\ data use the \href{https://www.govdata.de/dl-de/by-2-0}{German Data Licence--Attribution--Version 2.0}.
Under the PhysioNet Credentialed Health Data License 1.5.0, MIMIC-III v1.4 requires credentialing, human-subjects training, and a data-use agreement, so \MIMICReal\ and \MIMICSynth's MIMIC-based covariates cannot be regenerated without authorized access.
These terms do not establish redistribution rights for every derived benchmark artifact.

\paragraph{Models and artifacts.}
Apache-2.0 covers TabPFN-TS and Uni2TS code, Chronos-2 and TimesFM-2.5 code and checkpoints, and TabICL-v2's TabPFN-TS component; the TabPFN-v2 checkpoint uses Prior Labs License 1.1 (Apache-2.0 plus attribution), the rest of TabICL-v2 uses BSD-3-Clause, and Moirai-1.1-R-Large weights use CC BY-NC 4.0.
The five public baselines are available through the named repositories or model hubs.
The owner-authorized \ApolloPFN\ checkpoint remains closed and unavailable, so its predictions cannot be publicly regenerated; no code, data, model, or supplementary artifact accompanies this submission.

\paragraph{Participation and evaluation boundaries.}
Secondary, deidentified MIMIC-III data involve no participant recruitment, interaction, or intervention.
Pre-evaluation registries and frozen models use no benchmark-specific training or adaptation, reference retrieval, or test-outcome feedback.
Without a formal pretraining-contamination or deduplication audit, incompletely documented corpora may include related sources; novelty applies only to each evaluation history, not pretraining.

\newpage
\section*{NeurIPS Paper Checklist}

\begin{enumerate}

\item {\bf Claims}
    \item[] Question: Do the main claims made in the abstract and introduction accurately reflect the paper's contributions and scope?
    \item[] Answer: \answerYes{}
    \item[] Justification: The abstract and Sections~\ref{sec:introduction}--\ref{sec:discussion} state the benchmark scope, evaluation setting, findings, and limitations consistently with the reported results.

\item {\bf Limitations}
    \item[] Question: Does the paper discuss the limitations of the work performed by the authors?
    \item[] Answer: \answerYes{}
    \item[] Justification: Section~\ref{sec:experiments} notes that dataset characteristics vary jointly, and Appendix~\ref{app:dataset-registries} explains that scenario means pool separately generated or selected windows. Section~\ref{sec:discussion} identifies shared-history evaluation as future work. Appendix~\ref{app:access-reproducibility} documents access restrictions and possible pretraining overlap.

\pagebreak[3]
\item {\bf Theory assumptions and proofs}
    \item[] Question: For each theoretical result, does the paper provide the full set of assumptions and a complete (and correct) proof?
    \item[] Answer: \answerNA{}
    \item[] Justification: The paper presents no theoretical results or proofs.

    \item {\bf Experimental result reproducibility}
    \item[] Question: Does the paper fully disclose all the information needed to reproduce the main experimental results of the paper to the extent that it affects the main claims and/or conclusions of the paper (regardless of whether the code and data are provided or not)?
    \item[] Answer: \answerYes{}
    \item[] Justification: Appendices~\ref{app:metric-details}--\ref{app:access-reproducibility} document the metrics, dataset and scenario-subset construction, model settings, aggregation, and access limits. MIMIC-based reconstruction requires credentialed access, and the ApolloPFN checkpoint is not publicly available.

\item {\bf Open access to data and code}
    \item[] Question: Does the paper provide open access to the data and code, with sufficient instructions to faithfully reproduce the main experimental results, as described in supplemental material?
    \item[] Answer: \answerNo{}
    \item[] Justification: No code, data, model, or supplementary artifact is released with this submission due to the company policy. MIMIC-based data require credentialed access, and the evaluated ApolloPFN checkpoint is closed.

\item {\bf Experimental setting/details}
    \item[] Question: Does the paper specify all the training and test details (e.g., data splits, hyperparameters, how they were chosen, type of optimizer) necessary to understand the results?
    \item[] Answer: \answerYes{}
    \item[] Justification: Sections~\ref{sec:evaluation-framework}--\ref{sec:experiments} and Appendices~\ref{app:metric-details}--\ref{app:exp-setup} specify scenario construction, metrics, dataset generation or filtering, seeds, model settings, and aggregation.

\item {\bf Experiment statistical significance}
    \item[] Question: Does the paper report error bars suitably and correctly defined or other appropriate information about the statistical significance of the experiments?
    \item[] Answer: \answerNo{}
    \item[] Justification: The reported scores summarize one fixed benchmark realization (Appendix~\ref{app:exp-setup}). Sampling uncertainty and repeated-run variability remain unestimated.

\item {\bf Experiments compute resources}
    \item[] Question: For each experiment, does the paper provide sufficient information on the computer resources (type of compute workers, memory, time of execution) needed to reproduce the experiments?
    \item[] Answer: \answerYes{}
    \item[] Justification: Appendix~\ref{app:compute-resources} reports the AWS hardware and forward-pass inference time for every model--dataset--scenario evaluation, totaling $13{,}604.2$ seconds ($3.78$ hours), with stated exclusions.

\item {\bf Code of ethics}
    \item[] Question: Does the research conducted in the paper conform, in every respect, with the NeurIPS Code of Ethics \url{https://neurips.cc/public/EthicsGuidelines}?
    \item[] Answer: \answerYes{}
    \item[] Justification: The authors reviewed the NeurIPS Code of Ethics and confirm compliance. Data-access and reproducibility boundaries are documented in Appendix~\ref{app:access-reproducibility}.

\item {\bf Broader impacts}
    \item[] Question: Does the paper discuss both potential positive societal impacts and negative societal impacts of the work performed?
    \item[] Answer: \answerNA{}
    \item[] Justification: This preliminary work introduces an evaluation benchmark and does not deploy a decision-making system or release a model; we identify no direct societal impact requiring a dedicated discussion.

\item {\bf Safeguards}
    \item[] Question: Does the paper describe safeguards that have been put in place for responsible release of data or models that have a high risk for misuse (e.g., pre-trained language models, image generators, or scraped datasets)?
    \item[] Answer: \answerNA{}
    \item[] Justification: The submission releases no data or model requiring misuse safeguards; MIMIC-III is not redistributed and remains subject to PhysioNet access controls.

\item {\bf Licenses for existing assets}
    \item[] Question: Are the creators or original owners of assets (e.g., code, data, models), used in the paper, properly credited and are the license and terms of use explicitly mentioned and properly respected?
    \item[] Answer: \answerYes{}
    \item[] Justification: Appendices~\ref{app:dataset-registries}--\ref{app:access-reproducibility} credit the data, software, and model assets and state their licenses and access terms. The closed ApolloPFN checkpoint was used with authorization and is not redistributed.

\item {\bf New assets}
    \item[] Question: Are new assets introduced in the paper well documented and is the documentation provided alongside the assets?
    \item[] Answer: \answerNA{}
    \item[] Justification: No new code, dataset, model, or supplementary artifact is released with this submission.

\item {\bf Crowdsourcing and research with human subjects}
    \item[] Question: For crowdsourcing experiments and research with human subjects, does the paper include the full text of instructions given to participants and screenshots, if applicable, as well as details about compensation (if any)?
    \item[] Answer: \answerNA{}
    \item[] Justification: The study uses secondary analysis of deidentified MIMIC-III data and involves no crowdsourcing, participant recruitment, interaction, or intervention.

\item {\bf Institutional review board (IRB) approvals or equivalent for research with human subjects}
    \item[] Question: Does the paper describe potential risks incurred by study participants, whether such risks were disclosed to the subjects, and whether Institutional Review Board (IRB) approvals (or an equivalent approval/review based on the requirements of your country or institution) were obtained?
    \item[] Answer: \answerNA{}
    \item[] Justification: The study analyzes previously collected, deidentified MIMIC-III data under its access requirements and involves no participant recruitment, interaction, or intervention.

\item {\bf Declaration of LLM usage}
    \item[] Question: Does the paper describe the usage of LLMs if it is an important, original, or non-standard component of the core methods in this research? Note that if the LLM is used only for writing, editing, or formatting purposes and does \emph{not} impact the core methodology, scientific rigor, or originality of the research, declaration is not required.
    \item[] Answer: \answerYes{}
    \item[] Justification: LLM assistance was limited to code implementation of the author-defined scenario-subset registries and filtering pipeline, as disclosed in Appendix~\ref{app:exp-setup}.

\end{enumerate}

\end{document}